\documentclass[conference]{IEEEtran}

\usepackage{cite}

\usepackage[pdftex]{graphicx}
\graphicspath{{figures/}}
\DeclareGraphicsExtensions{.pdf,.png,.jpg}

\usepackage{amsmath,amssymb}
\usepackage{booktabs}
\usepackage[hidelinks]{hyperref}
\usepackage{balance}

\usepackage{hyperref}
\usepackage{amsmath}
\usepackage{amsthm}
\usepackage{graphicx}
\usepackage{textcomp}
\usepackage{xcolor}
\newtheorem{definition}{Definition}
\usepackage{comment}

\usepackage{adjustbox}
\usepackage{array}

\usepackage[ruled,linesnumbered]{algorithm2e}
\usepackage{algorithmicx}

\usepackage{multirow}
\usepackage{booktabs}
\usepackage{tabularx}
\usepackage{enumitem}
\usepackage{caption}
\usepackage{verbatim}

\usepackage{listings}
\usepackage{xcolor}

\usepackage{orcidlink}

\usepackage{tikz}

\newtheorem{theorem}{Theorem}[section]

\newtheorem{proposition}[theorem]{Proposition}
\newtheorem{example}[theorem]{Example}

\newcommand{\coolname}{$\textsc{TrashFuzz}$\xspace}

\begin{document}

\title{Natural Adversaries: Fuzzing Autonomous Vehicles with Realistic Roadside Object Placements}

\author{
\IEEEauthorblockN{
    Yang Sun\,\orcidlink{0000-0002-2409-2160}\hspace{20pt}
    Haoyu Wang\,\orcidlink{0009-0000-6379-5312}\hspace{20pt}
    Christopher M. Poskitt\,\orcidlink{0000-0002-9376-2471}\hspace{20pt}
    Jun Sun\,\orcidlink{0000-0002-3545-1392}
}
\IEEEauthorblockA{
    \textit{School of Computing and Information Systems}, \textit{Singapore Management University}, Singapore\\
    \{yangsun.2020, haoyu.wang.2024\}@phdcs.smu.edu.sg, \{cposkitt, junsun\}@smu.edu.sg
}
}

\maketitle

\begin{abstract}
    The emergence of Autonomous Vehicles (AVs) has spurred research into testing the resilience of their perception systems, i.e., ensuring that they are not susceptible to critical misjudgements. It is important that these systems are tested not only with respect to other vehicles on the road, but also with respect to objects placed on the roadside. Trash bins, billboards, and greenery are examples of such objects, typically positioned according to guidelines developed for the human visual system, which may not align perfectly with the needs of AVs. Existing tests, however, usually focus on adversarial objects with conspicuous shapes or patches, which are ultimately unrealistic due to their unnatural appearance and reliance on white-box knowledge. In this work, we introduce a black-box attack on AV perception systems that creates realistic adversarial scenarios (i.e., satisfying road design guidelines) by manipulating the positions of common roadside objects and without resorting to ``unnatural'' adversarial patches. In particular, we propose \coolname, a fuzzing algorithm that finds scenarios in which the placement of these objects leads to substantial AV misperceptions---such as mistaking a traffic light’s colour---with the overall goal of causing traffic-law violations. To ensure realism, these scenarios must satisfy several rules encoding regulatory guidelines governing the placement of objects on public streets. We implemented and evaluated these attacks on the Apollo autonomous driving system, finding that \coolname induced violations of 15 out of 24 traffic laws.
\end{abstract}


%
\IEEEpeerreviewmaketitle

\section{Introduction}
Autonomous Vehicles (AVs) are rapidly advancing, with several Level-4 systems successfully deployed in real traffic environments~\cite{sae2018taxonomy, waymo, apolloauto, TuSimple}. This progress has intensified interest in evaluating the resilience of their perception systems, which must reliably interpret not only vehicles and road obstacles but also roadside objects. Everyday items such as trash bins, billboards, and greenery are placed according to guidelines~\cite{cop,singaporestreetact2,SingaporeTreeGuidelines,HDS,KIMBLE,GRANGER} designed for human drivers, and may not align with the requirements of AV perception.

Most existing adversarial tests focus on conspicuous patches or irregularly shaped objects~\cite{cao2021invisible,hallyburton2022security,zhu2021can,sato2021dirty,cheng2022physical,muller2022physical}, such as MSF-ADV’s visibly altered traffic cone~\cite{cao2021invisible}. These attacks typically pursue objectives such as making an object “ignored’’ or mislocalised and often require white-box access to perception models. While such misperceptions may expose safety issues, their unrealistic appearance and impractical assumptions limit applicability on real roads.

These limitations motivate two questions.
First, can AV perception be deceived by natural, guideline-compliant roadside objects—those appearing entirely benign to human observers?
Second, if such misperceptions occur, can they escalate into serious safety issues, such as collisions or violations of traffic laws~\cite{Sun-Poskitt-et_al22a}? Demonstrating these rare but plausible scenarios in simulation is essential before AVs encounter them in the wild.

\begin{figure}[!t]
\includegraphics[width=1\linewidth]{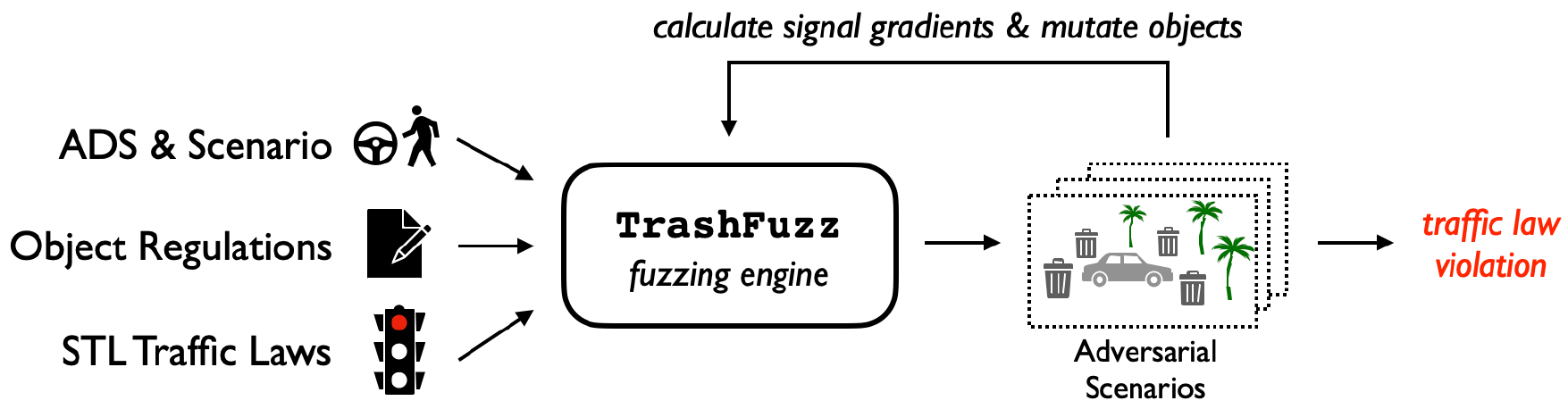}
\caption{\coolname workflow: mutating roadside-object placements to find natural scenarios that induce an autonomous vehicle to violate traffic laws}
\label{fig:workflow}
\end{figure}

To address these questions, we introduce \coolname, a black-box fuzzing approach that systematically searches for guideline-compliant object placements capable of misleading an Autonomous Driving System (ADS) into unsafe behaviour. \coolname generates scenarios using naturally shaped, patch-free objects (e.g., bins, benches, trees, hydrants), evaluates them in a high-fidelity simulator, and uses a greedy search to move “closer’’ to violating Signal Temporal Logic (STL) specifications of traffic laws. The fuzzer is constrained by regulatory documents—including the Code of Practice for Works on Public Streets~\cite{cop,singaporestreetact2}, greenery-provision guidelines~\cite{SingaporeTreeGuidelines}, and placement rules from major waste-management companies~\cite{HDS,KIMBLE,GRANGER}—ensuring all scenarios remain realistic. Figure~\ref{fig:workflow} illustrates the workflow.

Unlike LawBreaker~\cite{Sun-Poskitt-et_al22a}, which mutates the behaviour of agents (e.g., other vehicles or pedestrians) to induce traffic-law violations, \coolname treats regulation-compliant roadside object layouts themselves as the adversarial input space.
It introduces naturalness as a constraint, formalising real-world placement guidelines as executable rules that are enforced during scenario generation.
Within this constrained space, \coolname performs gradient-guided black-box optimisation over object types, positions, and orientations to minimise STL robustness and induce perception-driven law violations.

We implement \coolname for the Baidu Apollo ADS~\cite{apolloauto} with the Unity-based LGSVL simulator~\cite{rong2020lgsvl}, using the Unity Asset Store~\cite{UnityStore} to provide a wide range of everyday roadside objects. \coolname automatically enforces placement constraints and searches for scenarios that trigger STL-defined law violations. In our evaluation, \coolname efficiently generated scenarios that induced Apollo to violate 15 of 24 testable traffic laws. Most cases caused hesitation or failure to proceed; others induced aggressive behaviour, including one scenario in which an innocuous arrangement of bins overloaded the perception system and caused the ADS to misinterpret a red light as green.

These findings highlight that existing placement guidelines, designed primarily for human drivers, may be inadequate in the presence of AVs, and that testing pipelines must account for subtler, object-induced anomalies.

\section{Background and Problem}
\label{sec:back_problem}

In this section, we review the design of state-of-the-art ADS perception systems, define the attack goal and threat model, the DSLs for specifying safety properties, and highlight the key design challenges our solution must address.

\subsection{ADS Perception Systems}

State-of-the-art ADSs such as Apollo~\cite{latesetapollo}, Autoware~\cite{autoware}, and Waymo~\cite{waymo} share two key perception characteristics: multi-sensor integration and the use of Deep Neural Networks~(DNNs) for classification and segmentation.

\noindent \textbf{\emph{Integration of Multiple Sensors.}}
Modern Level-4 AVs typically fuse data from at least two sensor modalities—commonly cameras, LiDAR, and radar—to obtain a more accurate and robust view of the environment. While this multi-sensor setup enhances perception reliability, it does not guarantee resilience. For example, MSF-ADV~\cite{cao2021invisible} shows that carefully crafted objects can deceive both camera- and LiDAR-based detectors despite the presence of multiple sensors.

\noindent \textbf{\emph{DNN-based Classification and Segmentation.}}
DNNs are widely used in ADS perception pipelines for detecting and segmenting objects from camera and LiDAR data~\cite{latesetapollo,autoware}. However, DNNs are known to be vulnerable to adversarial manipulation. Prior work demonstrates that adversarial objects or patches can cause detectors to ignore obstacles~\cite{cao2021invisible}, misinterpret roadside billboards~\cite{zhou2020deepbillboard,patel2019adaptive,patel2021overriding}, deviate lane-centering controllers~\cite{sato2021dirty}, or corrupt depth estimation~\cite{cheng2022physical}. These examples are largely white-box attacks involving conspicuous or irregular adversarial patterns.

Defending DNNs against such attacks remains challenging. Although several defence mechanisms~\cite{kurakin2016adversarial,hinton2015distilling,papernot2016distillation,samangouei2018defense} show partial effectiveness, surveys~\cite{chakraborty2018adversarial,chakraborty2021survey} emphasise that no universal solution exists. Thus, systematic testing of ADS perception systems continues to be essential.

\subsection{Attack Goal and Threat Model}

\noindent \textbf{\emph{Attack Goal.}}
We propose a black-box adversarial attack on the perception systems of ADSs, with the goal of inducing behaviour in which the AV violates traffic laws.
This must be achieved by objects that follow conventional road design guidelines and appear benign to humans (i.e., no patches or unusual shapes).
Such scenarios are termed ``natural'': they are subsets of all possible scenarios and are characterised by restrictions on the placement of objects like bins, billboards, and greenery. We formulate the attack goal as follows:

\begin{definition}[Attack Goal]
\label{def:Attack_Goal}
    Let $\phi$ denote a user-specific specification of ADS behaviour, and $\varphi$ denote a formula specifying the rules of natural scenarios.
    Then, our attack goal is to find a scenario $\lambda$ such that $\lambda \vDash \varphi$ and there exists a trace $\pi$ of the ADS within $\lambda$ such that $\pi \nvDash \phi$.
\end{definition}
In this paper, the attack is divided into two tasks: (1)~generate a scenario $\lambda$ in which the placement of roadside objects satisfies some formalised regulatory rules $\varphi$; and (2)~optimise the placement of objects to induce a misperception that causes the ADS to violate a traffic law specification $\phi$.

\noindent \textbf{\emph{Threat Model.}}
Our attack assumes a black-box setting, in which the attacker has no access to the internal details of the perception system and its models.
The attacker is only able to observe the outputs of the perception system and chassis information of the ADS.
Furthermore, the attacker does not possess any knowledge of the inner architecture of the ADS or any algorithms that it utilises.
This ensures that the attack can be used to test the resilience of any ADS by treating its perception system as a black box.
The attacker's ability is limited to modifying and moving roadside objects.
They are further \emph{restricted to natural object placements} that satisfy government regulations~\cite{singaporestreetact} and company guidelines~\cite{HDS, KIMBLE, GRANGER} ensuring scenarios are realistic.
Furthermore, the attacker cannot utilise adversarial patches, as these appear as visually unnatural and require white-box knowledge of the perception models, which in our case the attacker does not have.

\subsection{Specifying Safety Properties}
\label{sec:Specification definition}

Our attack goal is to cause misperceptions that result in the violation of safety properties.
In the context of AVs, safety should not simply mean the absence of collisions, but also adherence to the rules of the road that drivers are supposed to abide by.
To this end, we adopt the property specification language used by LawBreaker~\cite{Sun-Poskitt-et_al22a}, as well as the project's existing specifications of the traffic laws of China and Singapore.
The specification language is based on STL formulas, and is evaluated with respect to traces of scenes, providing a way to automatically determine whether a tester-defined property was violated or not in a simulated run of the ADS.
We highlight the key features of the specification language below (the full syntax and semantics is given in~\cite{Sun-Poskitt-et_al22a}).

\begin{figure}[t]
    \centering\small
    {
    \begin{align*}
    \phi\;::=&\;\mu \;|\; \neg\phi\;|\; \phi_1 \lor \phi_2\;|\; \phi_1 \land \phi_2\;|\; \phi_1 \;\mathcal{U}_I\; \phi_2\\
    \mu\;::=&\;f(x_0, x_1, \ldots, x_k) \;\sim\; 0,\quad \sim \in \{>, \geq, <, \leq, =, \neq\}
    \end{align*}
    }
    \captionsetup{skip=0pt}
    \caption{Syntax of our specification language, a fragment of Signal Temporal Logic (STL). Here, $\phi$ denotes a formula, $\mu$ is an atomic predicate over real-valued functions $f$ of variables $x_i$, and $I = [l, u]$ is a bounded or unbounded time interval}
    \label{syntax}
\end{figure}

The high-level syntax of our specification language is formalised in Figure~\ref{syntax}. Temporal formulas $\phi$ are built from atomic predicates $\mu$ using Boolean connectives and the bounded temporal operator $\mathcal{U}_I$ (interpreted as `until over interval $I$'). A time interval $I$ is of the form $[l, u]$, where $l$ and $u$ are non-negative real-valued bounds. We adopt standard abbreviations: the \emph{eventually} operator $\Diamond_I \phi$ is syntactic sugar for $true \; \mathcal{U}_I \; \phi$, and the \emph{always} operator $\Box_I \phi$ abbreviates $\neg \Diamond_I \neg \phi$. When $I = [0, \infty]$, the interval may be omitted.

An atomic predicate $\mu$ is a relational constraint over a multivariate function $f(x_0, x_1, \ldots, x_k)$ and a constant threshold $0$, using standard relational operators. These variables $x_i$ are language-defined signals or features, such as vehicle speed, object distance, or perception confidence scores.

\begin{example}
    Suppose we have a signal variable $speed = \langle speed(0),  speed(1), \dots, speed(n) \rangle$, which represents the autonomous vehicle's speed throughout its journey.
    Then, we can create a simple Boolean expression $\mu = speed(t) < 100$ to check whether the speed of the vehicle is larger than $100km/h$, as illustrated in Fig~\ref{syntax}. 
    Note that $\mu$ can be regarded as a proposition of the form $100 - speed(t) > 0$ or $speed(t) - 100 < 0$.
    To verify whether $\mu$ holds true at all time steps, we can straightforwardly incorporate the temporal logic symbol ``always'', resulting in the formulation of $\varphi = \Box (speed < 100)$.
\end{example}

A specification is evaluated with respect to a trace $\pi$ of \emph{scenes}, denoted as $\pi=\langle \pi_0, \pi_1, \pi_2 \ldots, \pi_n \rangle$, where each scene $\pi_i$ is a valuation of the propositions at time step $i$ and $\pi_0$ reflects the state at the start of a simulation.
The language follows the standard semantics of STL (see e.g., \cite{maler2004monitoring}).

\subsection{Design Challenges}
\label{sec:DesignChallenges}

The goals above imply two key design challenges:

\noindent \textbf{\emph{C1. Restricting to natural scenarios.}}
A key objective of our work is to develop attack scenarios that appear as completely natural to humans.
Previous work, such as NADE~\cite{feng2021intelligent}, has focused on replicating natural driving behaviours of vehicles.
In contrast, our focus is on the strategic placement of typical roadside objects, including (but not limited to) trash bins, trees, and fire hydrants.
A significant challenge is knowing when the placement of such objects crosses the boundary between something innocuous to something suspicious.
To that end, we constrain the placement of objects according to existing regulations and guidelines (e.g., \cite{singaporestreetact,HDS, KIMBLE, GRANGER}), which we use to standardise the notion of natural scenarios.

In addition, by restricting our scenarios to natural ones, we impose constraints that significantly reduce the available space of modifications, requiring testing algorithms that navigate them more creatively.
We cannot arbitrarily modify the environment that the ADS is driving in.
For example, placing a trash bin directly in the middle of the road should be forbidden in our testing context, as this is clearly an unnatural scenario that violates standard guidelines on bin placement.

\noindent \textbf{\emph{C2. Large black-box search space.}}
Despite restricting the placement of objects to conventional road design guidelines, there remains an enormous number of possible combinations of placements, making the search space vast and challenging to navigate, particularly in a black-box setting.
For example, even when limiting to only 10 permissible placement locations, if each of them can be occupied by one of 10 different possible objects, we already have $10^{10}$ combinations of object placements.
Hence, an effective search algorithm is necessary.

\section{\coolname Methodology}
\label{sec:OurApproach}

\begin{figure}
    \centering
    \includegraphics[width=1\linewidth]{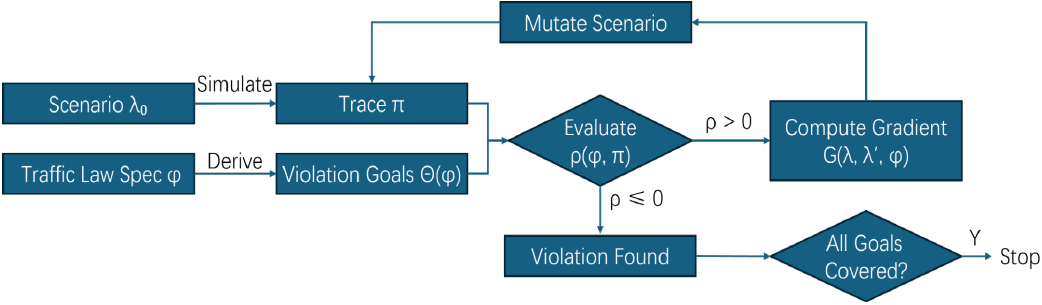}
    \caption{Overview of the \coolname approach}
    \label{fig:tf-overview}
\end{figure}

We propose \coolname to address these challenges.
Figure~\ref{fig:tf-overview} provides an overview, which we elaborate on below.

\subsection{Generating Natural Scenarios (C1)}
\label{sec:ourapporach_c1}

\begin{table}[t]
    \centering
    \caption{Specifications for natural scenarios, based on public road design guidelines}
    \label{tab:restriction_spec}
    \scriptsize
    \resizebox{\columnwidth}{!}{
    \begin{tabular}{c|p{0.85\linewidth}}
        $rule1$  & The fixed position attack objects should not be on the road or footway.\\
                 & $dis(obj.fix,road) \neq 0 \land dis(obj,footway) \neq 0$   \\ \hline

        $rule2$ & The fixed position attack objects (OG boxes) should be at least 0.6\,m away from the main road (2 lanes). \\
                & $lanenum(road) \ge 2 \implies dis(obj.fix,road) > 0.6$ \\ \hline

        $rule3$ & The fixed position attack objects cannot be near a minor road (1 lane). \\
                & $lanenum(road) < 2 \implies dis(obj.fix,road) > 10$\\ \hline

        $rule4$ & The splay corner of the entrance culvert, bin centre access, substation access, MDF room access, and fire engine access must be at least 1\,m from palms, 1.5\,m from small/medium trees, and 2.5\,m from large trees.\\
                & $(tree.h<1.5\implies dis(tree,corners)>1)\land(tree.h\ge1.5\land tree.h<5\implies dis(tree,corners)>1.5)\land(tree.h\ge5\implies dis(tree,corners)>2.5)$ \\ \hline

        $rule5$ & A scupper pipe/drain must be at least 1\,m from palms, 1.5\,m from small/medium trees, and 2.5\,m from large trees. \\
                & $(tree.h<1.5\implies dis(tree,pipe)>1)\land(tree.h\ge1.5\land tree.h<5\implies dis(tree,pipe)>1.5)\land(tree.h\ge5\implies dis(tree,pipe)>2.5)$\\ \hline

        $rule6$  & Trees must be at least 3\,m from a lamp post. \\
                & $dis(tree,lamp)>3$\\ \hline

        $rule7$  & Fixed objects (e.g., OG box, manhole, SCV box, lighting box, traffic light) must be at least 2\,m from small/medium trees, and 2.5\,m from large trees. \\
                & $(tree.h<5\implies dis(tree,obj.fix)>2)\land(tree.h\ge5\implies dis(tree,obj.fix)>2.5)$\\ \hline

        $rule8$  & Trees must be at least 3\,m from the edge of a footpath crossing. \\
                & $dis(tree,footpath)>3$\\ \hline

        $rule9$  & Trash bins should not block roads or footpaths. \\
                & $dis(bin,road)>0 \land (dis(bin,footpath)>0 \lor footpath.width-bin.width>1.5)$\\ \hline

        $rule10$ & Trash bins must be at least 0.5\,m from each other and from mailboxes, walls, lamp posts, utility boxes, vehicles, etc.\\ 
                & $dis(bin,obj)\ge0.5$ \\ \hline

        $rule11$ & Wheels/handles of bins must be oriented away from the road. \\
                & $bin.direction \times road.direction < 0.2$ \\ \hline

        $rule12$ & Ensure no object lies between trash bins and the road.\\
                & $dis(obj,road)+dis(bin,obj)>dis(bin,road)$     
    \end{tabular}}
\end{table}

As mentioned, we overcome C1 by utilising existing regulations and guidelines for the placement of various roadside objects.
In particular, we refer to government documents including the Code of Practice for Works on Public Streets~(COP)~\cite{cop} and Guidelines on Greenery Provision and Tree Conservation~(GGPTC)~\cite{SingaporeTreeGuidelines}.
The COP, formulated by Singapore's Land Transport Authority, delineates the necessary procedures and regulations for conducting activities on public roads, whereas the GGPTC, introduced by Singapore's National Parks Board, governs the placement and preservation of trees within urban landscapes.
For the placement of trash bins, we refer to the guidelines provided by three prominent recycling companies: HDS, KIMBLE, and GRANGER~\cite{HDS, KIMBLE, GRANGER}.

We formalise the notion of natural scenario by translating the regulations/guidelines described in these documents into mathematical formulas.
Though in natural language, the guidelines are quite specific, allowing for a relatively direct translation of the rules to formulas.
Table~\ref{tab:restriction_spec} lists these 12 specifications that jointly define natural scenarios, all of which must be satisfied.
Specifications 1--3 refer to section 6.5 of the COP~\cite{cop}.
Specifications 4--8 refer to pages 95-97 of the GGPTC~\cite{SingaporeTreeGuidelines}.
Specifications 9--11 pertain to key aspects of the guidelines provided by the waste recycling companies~\cite{HDS, KIMBLE, GRANGER}.
Note that while we have focused on guidelines from one country to ensure consistency between our rules, the general approach of formalising rules can be applied to the guidelines of any other country or jurisdiction.

In our rules, the function $dis(a, b)$ returns the distance between two objects, $a$ and $b$, while the function $lanenum(road)$ returns the count of lanes within the designated $road$. 
Each distinct category of objects is denoted by a class variable.
For example, the class $tree$ serves as a general representation for trees and includes a characteristic attribute, $tree.h$, corresponding to height.

It is crucial to have a diverse range of attack objects at our disposal.
Several simulators, including AWSIM~\cite{AWSIM} and LGSVL~\cite{rong2020lgsvl}, are built upon the Unity Real-Time Development Platform~\cite{UnityEngine}. 
The Unity Asset Store~\cite{UnityStore} has a vast library of 3D asset packages, with an open-door policy that allows anyone to contribute their creations.
Each asset package provides necessary elements, such as meshes of 3D objects and 2D textures, for constructing objects inside the simulators. 
We can access thousands of objects that are commonly encountered during driving such as rocks, trash bins, and hydrants for free.
Hence, our strategy involves exploring Unity's asset library~\cite{UnityStore} to discover suitable adversarial objects.

In this work, we leverage a variety of freely accessible 3D asset packages from the Unity Asset Store~\cite{UnityStore} to integrate commonly encountered roadside objects.
These include trash bins in a variety of shapes and colours, benches constructed from different materials and exhibiting various shapes, a wide array of tree types, hydrants, and trash bags.
Crucially, these objects can be effortlessly spawned within Unity-based simulators, such as LGSVL~\cite{rong2020lgsvl}.
We provide all these assets on our website~\cite{ourweb}, offering a convenient way for users to integrate them into the LGSVL simulator environment directly.

\subsection{Greedy Gradient-Guided Black-Box Search (C2)}
\label{sec:greedy_guided_algorithm}

We require an efficient algorithm for identifying attack scenarios in a black-box setting.
Given our objective is to induce ADS behaviours that violate traffic law specifications, it is important that the algorithm can uncover \emph{different} ways to violate the specification.
For instance, consider the specification $(speed > 10) \wedge (speed < 100)$.
There are two distinct ways to violate it: first by inducing $speed$ to a value smaller than 10, and second by inducing it to a value larger than 100.

In this section, we present the detailed design of our algorithm.
Given the time-consuming nature of simulator-based evaluations for ADSs, our primary goal is to attain convergence as fast as possible. To achieve this, our algorithm is \emph{greedy}, in the sense of initially exploring adversarial scenarios that appear most likely to violate the given specification, and is also \emph{gradient guided}, in the sense that it systematically expands its exploration based on these initial findings.

We present the algorithm in three parts.
First, we delve into the encoding of (natural) scenarios as vectors, which forms the foundation of subsequent steps (e.g., mutation).
Second, we present the design of our objective function which provides a quantified way to sufficiently cover different ways of violating a specification, ensuring diversity in the results.
Finally, we introduce our new greedy gradient-guided search algorithm, which forms the core of our approach.

\subsubsection{Encoding of Scenarios}
\label{sec:ScenarioEncoding}

We represent each adversarial scenario as a vector $\phi$ with dimensions $n \times d$.
Here, $n$ refers to the number of adversarial objects, while $d$ corresponds to the four essential dimensions of each adversarial object: \emph{forward}, \emph{right}, \emph{rotation}, and \emph{type}.
The \emph{forward} and \emph{right} dimensions indicate the relative position of the adversarial object with respect to the start position of the ego vehicle.
For example, if the value for \emph{forward} is $10$, and the value for \emph{right} is $-2$, the object is located at the left front of the ego vehicle.
The \emph{rotation} dimension represents the placement angle of the adversarial object.
The \emph{type} is a sequence number corresponding to the adversarial objects library, allowing us to easily identify and categorise different types of adversarial objects.
In addition, each scenario also provides the initial and destination positions for the ADS journey.

\begin{table}[t]
    \centering
    \caption{An example of an encoded scenario}
    \vspace{-5pt}
    \label{tab:example_of_encoded_scenario}

    \resizebox{\linewidth}{!}{%
    \begin{tabular}{|c|c|c|c|c|} \hline
         Adversarial Objects & forward & right & rotation & type \\
         \hline
         $\mathtt{obstacle1}$ & 25.04 & 2.63 & 25.78$^\circ$ & Bench1  \\
         $\mathtt{obstacle2}$ & 49.45 & 6.23 & 9.21$^\circ$ & BigTrashBin  \\
         $\mathtt{obstacle3}$ & 60.55 & 4.14 & 91.98$^\circ$  & TrashBin(Yellow)  \\
         $\mathtt{obstacle4}$ & 57.85 & 7.98 & 49.89$^\circ$  & TrashBin(Red) \\
         $\mathtt{obstacle5}$ & 18.95 & -9.83 & 290.48$^\circ$ & TrashBin(Green) \\
         \hline
    \end{tabular}
    }
    \vspace{-5pt}
\end{table}

Table~\ref{tab:example_of_encoded_scenario} provides an example scenario encoding.
Within this scenario, there are five adversarial objects.
Four of these objects, namely a bench, a large trash bin, a yellow trash bin, and a red trash bin, are strategically placed to the front-right of the ego vehicle's starting point, each with distinct rotation angles (designated as $obstacle1\dots4$).
Additionally, a green trash bin is positioned in the front-left with a rotation angle of 290.48 degrees, labelled as $obstacle5$.

\subsubsection{Coverage-based Objective Function Design}
\label{sec:coverage-based-objective}
We treat the attack generation process as an optimisation problem.
\begin{definition}
Let $\Theta$ be a set of traffic law specifications and $\Lambda$ a set of adversarial scenarios. Our optimisation problem is:
{\begin{align*}
    Minimise: \sum_{\phi \in \Theta} \min \limits_{\lambda \in \Lambda}\{\rho(\phi, \pi_\lambda)\}
\end{align*}}

\noindent where $\pi_\lambda$ is the trace of ADS in the face of an adversarial scenario $\lambda$ and function $\rho(\phi, \pi)$ is the robustness value of trace $\pi$ evaluated under specification $\phi$.
\end{definition}

We thus require a set of adversarial scenarios that minimises the robustness function across several property specifications. 
Intuitively, the robustness function reflects the \emph{distance} to a specification violation, meaning it measures how close the current trace $\pi$ is to violating the specification.
First, we must establish a precise definition for the fundamental \emph{objective function}, denoted as $\rho(\phi, \pi)$.
Second, we require a systematic approach to replace the specification with a set of smaller formulas $(\Theta)$, each one characterising a different way to violate the original specification.


We define our objective function using the quantitative semantics of STL~\cite{maler2004monitoring, deshmukh2017robust,nivckovic2020rtamt}, which produces a numerical \emph{robustness} degree. 
\begin{definition}[Quantitative semantics]\label{def:Quantitative_Semantics}
Given a trace $\pi$ and a formula $\phi$, the \emph{quantitative semantics} is defined as the robustness degree $\rho(\phi, \pi,t)$, computed as follows.
Recall that propositions $\mu$ are of the form $f(x_0,x_1,\cdots,x_k) \sim 0$.
\begin{equation*}
  \rho(\mu, \pi, t) =
    \begin{cases}
      -\pi_t(f(x_0,x_1,\cdots, x_k)) & \text{if $\sim$ is $\leq$ or $<$}\\
      \pi_t(f(x_0,x_1,\cdots, x_k)) & \text{if $\sim$ is $\geq$ or $>$}\\
      \mid \pi_t(f(x_0,x_1,\cdots, x_k)) \mid & \text{if $\sim$ is $\neq$}\\
      -\mid \pi_t(f(x_0,x_1,\cdots, x_k)) \mid & \text{if $\sim$ is $=$}
    \end{cases}       
\end{equation*}
where $t$ is the time step and $\pi_t(e)$ is the valuation of expression $e$ at time $t$ in $\pi$;
{\small
\begin{align*}
\rho(\neg\phi,\pi,t) & = -\rho(\phi,\pi,t) \\
\rho(\phi_1 \land \phi_2,\pi,t) & = \min\{\rho(\phi_1,\pi,t),\rho(\phi_2,\pi,t)\} \\
\rho(\phi_1 \lor \phi_2,\pi,t) & = \max\{\rho(\phi_1,\pi,t),\rho(\phi_2,\pi,t)\} \\
\rho(\phi_1 \;\mathtt{U_I}\; \phi_2,\pi,t) & = \sup_{t_1 \in t+\mathtt{I}} \min \{\rho(\phi_2,\pi,t_1), \inf_{t_2 \in [t,t_1]} \rho(\phi_1,\pi,t_2)\}
\end{align*}
}
where $t+I$ is the interval $[l+t,u+t]$ given $I=[l,u]$. 
\qed
\end{definition}

Intuitively, this quantitative semantics computes how `close' the ADS is from violating the given property specification $\phi$. 
Note that the smaller $\rho(\phi,\pi,t)$ is, the closer $\pi$ is to violating $\phi$.
If $\rho(\phi,\pi,t) \leq 0$, this means $\phi$ is violated.
We write $\rho(\phi,\pi)$ to denote $\rho(\phi,\pi,0)$; $\pi \vDash \phi$ to denote $\rho(\phi,\pi,t) > 0$; and $\pi \not \vDash \phi$ to denote $\rho(\phi,\pi,t) \leq 0$. Note that time is discrete in our setting. 

\begin{example}
\label{example:robustness calculation}
Let $\varphi = \Box (speed < 100)$, i.e., the vehicle's speed must always remain below $100$ km/h.
Suppose $\pi$ is $\langle (speed \mapsto 0, \dots), (speed \mapsto 0.5, \dots), \cdots (speed \mapsto 90, \dots) \rangle$, where the ego vehicle's max $speed$ is $90$km/h at the last time step.
We have $\rho(\varphi, \pi) = \rho(\varphi, \pi, 0) = min_{t \in [0, |\pi|]} ( 100 - \pi_t(speed) ) = 10$.
This means that trace $\pi$ satisfies $\varphi$, and the robustness value is 10. 
\end{example}

Next, we transform a specification into a set of smaller formulas, such that the violation of any one formula implies the \emph{violation of the original specification}.
The rationale is that this will allow us to evaluate the robustness degree with respect to multiple different ways of violating the original specification, rather than just focusing on the `easiest' one.
Here, we adopt the methodology of LawBreaker~\cite{Sun-Poskitt-et_al22a}, which defines a method to derive a set of violation goals based on the given complex specification $\phi$.
We write $\Theta(\phi)$ to denote a set of `violation goals' of $\phi$, i.e., that satisfy the following proposition:
\begin{proposition}
\label{prop:subdivision}
    For STL formulas $\phi$ and traces $\pi$:
    \begin{align*}
        \forall \varphi \in \Theta(\phi).~\pi \nvDash \varphi \implies \pi \nvDash \phi
    \end{align*} 
\end{proposition}

The set of violation goals $\Theta(\phi)$ is calculated as follows:
\begin{align*}
 & \Theta(\mu) = \{\mu\}; \Theta(\lnot \phi) = \Theta(N(\phi));\\
 & \Theta(\phi_1 \land \phi_2) = \Theta(\phi_1) \cup \Theta(\phi_2); \Theta(\bigcirc \phi) = \{\bigcirc x ~|~ x \in \Theta(\phi) \};\\
 & \Theta(\phi_1 \ \mathcal{U_I} \ \phi_2) = \{x \ \mathcal{U_I}\ y ~|~ x \in \Theta(\phi_1) \land y \in \Theta(\phi_2)\};
\end{align*}
Here, $\mu$ is a Boolean expression and function $N(\phi)$ returns the negation of the formula $\phi$ without operator $\lnot$.
This eliminates the need for an additional calculation to determine $\Theta(\lnot \phi)$, as all formulas containing the operator $\lnot$ have already been transformed into their equivalent formulas without operator $\lnot$.
\begin{proposition}
\label{prop:negation_of_a_spec}
For STL formulas $\phi$ and traces $\pi$:
\vspace{-4pt}\begin{align*}
    \forall \phi.~\forall \pi.~\rho(N(\phi), \pi) = \rho(\lnot \phi, \pi).
\end{align*}
\end{proposition}
\begin{definition}[Specification Negation]\label{def:STL_Negation}
Given a specification formula $\phi$, the negation of this formula, $N(\phi)$, is defined:
\begin{align*}
& N(\mu) = n(\mu); N(\lnot \phi) =\phi; N(\phi_1 \land \phi_2) = N(\phi_1) \lor N(\phi_2); \\
& N(\phi_1 \lor \phi_2) =  N(\phi_1) \land N(\phi_2); N(\bigcirc \phi) = \bigcirc N(\phi);\\
& N(\phi_1 \;\mathtt{U_I}\; \phi_2) = \left(\Box_\mathtt{I}  N(\phi_2) \right) \lor \left( N(\phi_2) \;\mathtt{U_I}\; ( N(\phi_1) \land N(\phi_2) ) \right); 
\end{align*}
\noindent where $n(\mu)$ is defined as follows:
\begin{equation*}
  n(\mu) =
    \begin{cases}
    e_1 \geq(resp. >) e_2  & \text{if $\mu$ is $e_1 <(resp. \leq) e_2$}; \\
    e_1 \leq(resp. <) e_2  & \text{if $\mu$ is $e_1 >(resp. \geq) e_2$}\\
    e_1 \neq(resp. =) e_2  & \text{if $\mu$ is $e_1 =(resp. \neq) e_2$}\\
    \end{cases}        
\end{equation*}
\end{definition}

Propositions~\ref{prop:subdivision} and~\ref{prop:negation_of_a_spec} can be proved by structural induction. We refer the readers to~\cite{Sun-Poskitt-et_al22a} for the detailed proofs.

\begin{example}
The value of $\Theta(\phi)$ is calculated recursively.
For example, given a specification $\phi = \Box (\mu_1 \land \mu_2)$, we can obtain $\Theta(\phi)$ as follows:
\begin{enumerate}
    \item First, we apply $\Theta$ to the primitive elements:
    $\Theta(\mu_1) = \{\mu_1\}, \Theta(\mu_2) = \{\mu_2\}$
    \item Then we have:
    $\Theta(\mu_1 \land \mu_2) = \Theta(\mu_1) \cup \Theta(\mu_2) = \{\mu_1, \mu_2\}$
    \item Given $\Theta(\mu_1 \land \mu_2)$, we can get: \\
    $\Theta(\Box (\mu_1 \land \mu_2)) = \{\Box (\mu_1), \Box (\mu_2) \} $
    \item The final result is $\Theta(\phi) =\{\Box (\mu_1), \Box (\mu_2) \}$.
\end{enumerate}
\noindent i.e., there exist two violation goals for $\phi$, and the violation of either one of them can result in the violation of $\phi$.
\end{example}

\subsubsection{Greedy Gradient Guided Query}
With the derivation of violation goals reviewed, we now propose our novel greedy gradient-guided query algorithm.
We first describe the gradient calculation process before presenting the algorithm as a whole.

In our black-box setting, gradient calculations are important for guiding the generation of adversarial scenarios.
These gradients are constructed by differentiating between the original scenario and the scenario resulting from modifications.
We define the gradient calculation below:
\begin{definition}
Let $\phi$ be a property specification, $\lambda$ be the initial scenario, $\lambda'$ be the scenario after modification, and $E(\lambda)$/$E(\lambda')$ denote the encoding of scenario $\lambda$/$\lambda'$.
The gradient calculation is formally defined as follows:
\begin{align*}
G(\lambda, \lambda', \phi) = \frac{\rho(\phi, \pi) - \rho(\phi, \pi')}{E(\lambda) - E(\lambda')}
\end{align*}
where $\pi$ and $\pi'$ are the ADS traces under scenarios $\lambda$ and $\lambda'$, respectively.
\end{definition}
Intuitively, the gradient calculation reveals the extent to which modifying the scenario influences the behaviour of the ADS.
A positive gradient value signifies that the current modification enhances robustness with respect to the property specification, while a negative gradient value suggests a decrease in robustness due to the modification.
Furthermore, the magnitude of the gradient value reflects the magnitude of the modification's impact on the robustness value.

\begin{algorithm}[t]
{\footnotesize
\caption{The \coolname Algorithm}\label{alg:overall-algorithm}
\KwIn{$\phi$ (traffic law), $n$ (number of adversarial objects), and $M$ (maximal queries)}
\KwOut{A test suite $\Gamma$}
Let $\Theta_r = \Theta(\phi)$ be the set of uncovered formulas in $\Theta(\phi)$\;
Let $seed$ be a pair storing a gradient value and mutable element; initially the gradient value $seed.gradient$ is 0 and the mutable element $seed.element$ is $Null$\;
Let $\lambda_0$ be a randomly generated adversarial scenario\;
Let $\Gamma$ be an empty set\;
\While{$\Theta_r$ is not empty and not timeout}
{
    Execute $\lambda_0$ via simulation and obtain trace $\pi_0$\;
    Mutate each adversarial object in $\lambda_0$ and get a few adversarial scenarios $\lambda_1, \dots, \lambda_n$\;
    Execute $\lambda_1, \dots, \lambda_n$ via simulation and obtain trace $\pi_1, \dots, \pi_n$\;
    \For{each $\pi_k$ in $\pi_1, \dots, \pi_n$ }{
        \For{each $\varphi$ in $\Theta_r$ }{
            \If {$\rho(\varphi, \pi_k) \leq 0$}{
                Remove $\varphi$ from $\Theta_r$; Add $s_k$ into $\Gamma$\;
            }
            \ElseIf{$\vert G(\lambda_0, \lambda_k, \varphi) \vert > \vert seed.gradient \vert$ }{
                 $seed.gradient = G(\lambda_0, \lambda_k, \varphi)$\;
                 Store the mutated element of $s_k$ to $seed.element$\;
            }      
        }
    }
    Generate the next $s_0$ based on $seed$;
}
\Return $\Gamma$
}
\end{algorithm}

The overall \coolname~algorithm is shown in Algorithm~\ref{alg:overall-algorithm}.
First, we generate a random scenario as the initial test case and decompose the traffic law specification $\phi$ to obtain $\Theta_r$, then initialise $seed$ with $seed.gradient$ value $0$ and $seed.element$ value $Null$. 
Note that $seed$ is a pair storing a gradient value and the corresponding mutable element.

Next, using the initial adversarial scenario $\lambda_0$, we execute it to generate the trace $\pi_0$.
The adversarial scenario contains several adversarial objects, and for each of these objects, we apply random mutations to create new scenarios. Note that the mutations have been designed to ensure that the ADS remains compliant with the rules listed in Table~\ref{tab:restriction_spec}, as well as with other physical constraints such as the requirement for objects to be grounded.
For each newly generated scenario, we also execute it to produce a trace and assess it against all the remaining specifications in set $\Theta_r$.
We remove $\varphi$ from the set if it is violated, i.e., if $\rho(\varphi, \pi_i) \leq 0$, and add the corresponding scenario $\lambda_i$ to the output set $\Gamma$.
After executing and processing all the cases generated by the initial test case, we utilise the gradient calculation defined above to choose the seed for the next round.
In essence, we designate the case with the largest absolute gradient value as the seed for the subsequent round. 

This process is repeated until all the elements in set $\Theta(\phi)$ are covered or the maximal number of generations $M$ (as defined by the user) has been reached.

\section{Implementation and Evaluation}
\label{sec:evaluation}
In this section, we present our implementation of \coolname for the Apollo ADS and evaluate its effectiveness in identifying natural scenarios that lead to traffic law violations, as well as other key aspects. For additional experimental materials and results, we direct readers to our website~\cite{ourweb}.

\subsection{Implementation}

Implementing \coolname involves three broad steps.
First, some basic integration with and between the existing ADS and simulator, including a communication bridge, a trace generator, and the means of both checking a specification as well as calculating its robustness with respect to a trace.
Second, the generation of natural scenarios consisting of (benign-appearing) adversarial roadside objects, as detailed in Section~\ref{sec:ourapporach_c1}.
Finally, the implementation of our greedy gradient-based fuzzing engine, as elaborated on in Section~\ref{sec:greedy_guided_algorithm}.

The target of our evaluation is the LGSVL~\cite{rong2020lgsvl} simulator with version 7.0 of the Apollo ADS~\cite{apolloauto} (the latest version stable with LGSVL).
To establish communication between Apollo and LGSVL, we utilise the official communication bridge, CyberRT~\cite{CyberRT}.
For trace generation and specification evaluation/robustness, we use the implementation strategy described by LawBreaker~\cite{Sun-Poskitt-et_al22a}.
In particular, the trace generator creates a trace $\pi$ by processing messages subscribed to by the ADS, whereas for specifying traffic laws, we utilise the domain-specific language provided by LawBreaker.
To implement the quantitative semantics of those traffic laws, we embedded the \emph{RTAMT} tool~\cite{nivckovic2020rtamt} to compute the robustness of the specifications with respect to the trace obtained from the bridge.
(Note that implementing \coolname for Autoware/CARLA would follow a largely similar strategy, except for substituting the ROS bridge~\cite{carlaautowarebridge} for CyberRT.)

In order to generate natural scenarios, we sourced common roadside objects from the Unity Asset Store~\cite{UnityStore}.
As LGSVL is a Unity-based simulator, these could be directly placed into scenarios.
The roadside objects we utilised in \coolname have been assembled into a single library that we have made available online~\cite{ourweb}.
To ensure the objects are placed according to conventional road design guidelines, we constrained their placement according to the functions in Table~\ref{tab:restriction_spec}.
These functions take a scenario as input and return a Boolean value to indicate whether the generated scenario is valid or not according to the guidelines.

Finally, we integrated the above components and implemented our greedy gradient-based fuzzing engine as described in Section~\ref{sec:greedy_guided_algorithm}.
Our experiments were conducted using two machines, each equipped with 32GB of memory, an Intel i7-10700k CPU, and an RTX 2080Ti graphics card.
These two machines run on Linux (Ubuntu 20.04.5 LTS) and Windows (Windows 10 Pro) operating systems, respectively.

\subsection{Evaluation}
We conducted experiments to answer four Research Questions~(RQs):

\begin{itemize}
   \item \noindent \textbf{RQ1}: Can \coolname produce valid scenarios?
   \item \noindent \textbf{RQ2}: Can natural attack scenarios cause traffic law violations?
   \item \noindent \textbf{RQ3}: Can our greedy gradient-based search algorithm cover more violations than the most comparable fuzzer?
   \item \noindent \textbf{RQ4}: Do other ADS perception stacks exhibit similar object-induced misclassifications?
\end{itemize}

\begin{table}[]
\scriptsize
\centering
\captionsetup{skip=5pt}
\caption{Percentage of valid scenarios}
\label{tab:natural-scenario-percent}

\resizebox{\linewidth}{!}{%
{\large
\begin{tabular}{|c|c|*{7}{c|}}
\hline
\multicolumn{2}{|c|}{Atk Obj}          & 1         & 2         & 3         & 4         & 5         & 6         & 7  \\\hline

\multirow{8}{*}{Rand} 
& \text{rule1}   & 79.9 & 63.7 & 50.1 & 39.9 & 32.5 & 25.2 & 20.3 \\ \cline{2-9} 
& \text{rule2-3} & 77.7 & 60.2 & 46.2 & 35.2 & 28.3 & 21.1 & 16.4 \\ \cline{2-9} 
& \text{rule4-6} & 91.1 & 83.5 & 76.3 & 69.0 & 63.6 & 57.4 & 53.2 \\ \cline{2-9} 
& \text{rule9}   & 85.1 & 73.7 & 62.9 & 53.9 & 45.8 & 38.3 & 34.1 \\ \cline{2-9} 
& \text{rule10}  & 100  & 99.99 & 99.98 & 99.99 & 99.97 & 99.96 & 99.95 \\ \cline{2-9} 
& \text{rule11}  & 89.2 & 80.7 & 71.5 & 64.9 & 58.0 & 51.8 & 45.8 \\ \cline{2-9} 
& \text{rule12}  & 100  & 100   & 100   & 100   & 100   & 100   & 100 \\ \cline{2-9} 
& \textbf{all}   & \textbf{56.0} & \textbf{32.0} & \textbf{17.6} & \textbf{10.1} & \textbf{5.8} & \textbf{3.1} & \textbf{1.6} \\ \hline

\text{Ours} & \textbf{all} & \textbf{100} & \textbf{100} & \textbf{100} & \textbf{100} & \textbf{100} & \textbf{100} & \textbf{100} \\ \hline
\end{tabular}
}%
} 

\vspace{-5pt}
\end{table}

\noindent \textbf{\emph{RQ1: Can \coolname produce valid scenarios?}}
Given the specifications shown in Table~\ref{tab:restriction_spec}, we first investigate whether \coolname can effectively generate valid scenarios that satisfy them.
We compare \coolname with randomly generated scenarios.
Our evaluation involved analyzing 10,000 randomly generated scenarios, each with a fixed number of attack objects. Note that in these scenarios, we randomly place attack objects within a designated area, i.e., a 150m $\times$ 60m space in front of the AV. 
The results are shown in Table~\ref{tab:natural-scenario-percent}. Here, the numbers~(1 to 7) in the first row indicate the number of attack objects. We present the percentage of scenarios that satisfy the given specification. A higher percentage value indicates a greater number of valid scenarios.
Rule2 and rule3 have been merged into one specification concerning the distance of fixed-position objects from the road. Likewise, rule4--6 are combined into one specification that evaluates the positioning of trees.
This analysis demonstrates that a majority of the randomly generated scenarios are invalid with respect to public road design guidelines. Moreover, as the number of attack objects increases, the likelihood of scenarios deviating from these guidelines rises. 
This result indicates that most randomly generated scenarios are not natural. For instance, these objects are often placed directly on the road, a violation of $\mathtt{rule1}$ or $\mathtt{rule9}$ as outlined in Table~\ref{tab:restriction_spec}.
As shown by the final row of Table~\ref{tab:natural-scenario-percent}, \coolname generates scenarios that satisfy the placement rules by construction (invalid mutations are discarded via lightweight constraint checks), yielding a much higher proportion of natural scenarios than random placement.

\begin{table}[]
\scriptsize
\centering
\captionsetup{skip=5pt}
\caption{Violations of traffic laws; note that \coolname operates under natural scenario restrictions}
\label{tab:coverage_of_failures}

\resizebox{\linewidth}{!}{%
{\footnotesize
\begin{tabular}{|c|c|c|c|c|}
\hline
\multicolumn{2}{|c|}{\textbf{Traffic Laws}}  & \textbf{LawBreaker} & \textbf{\coolname} & \textbf{Law Concerns} \\ \cline{1-5}

\multirow{3}{*}{Law38}
    & sub1 & $\surd$ & $\surd$ & green light \\ \cline{2-5}
    & sub2 & $\surd$ & $\surd$ & yellow light \\ \cline{2-5}
    & sub3 & $\surd$ & $\surd$ & red light \\ \hline

\multicolumn{2}{|c|}{Law44} & $\surd$ & $\surd$ & lane change \\ \hline

\multirow{2}{*}{Law45}
    & sub1 & $\times$ & $\times$ & speed limit \\ \cline{2-5}
    & sub2 & $\times$ & $\times$ & speed limit \\ \hline

\multirow{2}{*}{Law46}
    & sub2 & $\surd$ & $\surd$ & speed limit \\ \cline{2-5}
    & sub3 & $\surd$ & $\surd$ & speed limit \\ \hline

\multicolumn{2}{|c|}{Law47} & $\surd$ & $\surd$ & overtake \\ \hline
\multicolumn{2}{|c|}{Law50} & $\times$ & $\times$ & reverse \\ \hline

\multirow{6}{*}{Law51}
    & sub3 & $\surd$ & $\surd$ & traffic light \\ \cline{2-5}
    & sub4 & $\surd$ & $\surd$ & traffic light \\ \cline{2-5}
    & sub5 & $\surd$ & $\surd$ & traffic light \\ \cline{2-5}
    & sub6 & $\times$ & $\times$ & traffic light \\ \cline{2-5}
    & sub7 & $\times$ & $\surd$ & traffic light \\ \hline

\multicolumn{2}{|c|}{Law52 sub2-4} & $\times$ & $\times$ & priority \\ \hline
\multicolumn{2}{|c|}{Law53}         & $\times$ & $\times$ & traffic jam \\ \hline

\multirow{2}{*}{Law57}
    & sub1 & $\surd$ & $\surd$ & left turn signal \\ \cline{2-5}
    & sub2 & $\surd$ & $\surd$ & right turn signal \\ \hline

\multicolumn{2}{|c|}{Law58} & $\surd$ & $\surd$ & warning signal \\ \hline
\multicolumn{2}{|c|}{Law59} & $\surd$ & $\surd$ & signals \\ \hline

\multirow{1}{*}{Law62}
    & sub8 & $\times$ & $\times$ & honk \\ \hline
\end{tabular}
}%
} 

\end{table}

\noindent \textbf{\emph{RQ2: Can natural attack scenarios cause traffic law violations?}}
To answer this question, we applied our fuzzing algorithm to systematically test all testable traffic law specifications (including the multiple sub-laws) provided with the LawBreaker~\cite{Sun-Poskitt-et_al22a} tool.
Our experiment covered all `testable' laws, excluding those that cannot be tested due to limitations in the simulator.
For instance, traffic regulations related to traffic lights with arrow lights are non-testable because LGSVL only supports traffic lights with circular lights. 

The results are summarised in Table~\ref{tab:coverage_of_failures}. Here, we present all the testable laws applicable to AVs within the platform. However, certain other relevant laws cannot be tested due to limitations in the available maps. For example, law 49 specifies that vehicles should not make U-turns at railway crossings, sharp bends, steep slopes, or tunnels. Unfortunately, the current maps of existing platforms~(LGSVL and CARLA) do not include representations of these specific locations, rendering them untestable under the current conditions.

Within this table, the `LawBreaker' and `\coolname' columns indicate whether Apollo violated the specific traffic regulation listed in the first two columns.
We mark $\surd$ for a traffic law $\phi$ if and only if $\pi \nvDash \phi$ happened for a scenario generated by LawBreaker or \coolname.
We repeated each test case at least three times to ensure reproducibility and mitigate potential flakiness in simulation-based ADS testing~\cite{Osikowicz-et_al25a}.
Note that the tools are violating laws using different mutation strategies: LawBreaker mutates the driving behaviour of other vehicles and pedestrians, whereas \coolname mutates only the placement of roadside objects subject to the roadside design guidelines (Table~\ref{tab:restriction_spec}).
\coolname's strategy is thus much more challenging due to the additional restrictions.

As can be seen from the table, even while constrained to satisfy road design guidelines, \coolname still manages to incite violations of a majority of the traffic laws.
In comparison to LawBreaker, \coolname covers all the traffic laws that LawBreaker violates.
Additionally, \coolname has the capability to activate an additional sub-law, specifically sub-law 7 of Article 51,
which regulates that turning vehicles yield to straight-moving vehicles, whereas pedestrians and right-turning vehicles travelling in the opposite direction yield to left-turning vehicles. The activation of this supplementary sub-law is triggered by the presence of adversarial objects along the roadside. Note that \coolname found an additional law violation compared to LawBreaker \emph{despite the restriction} to modifying roadside object placements only.

We identified two broad categories of issues among the tests generated by \coolname, which we elaborate on below. 
These misjudgments stem from flaws in the perception system of the ADS. While the camera images and lidar points received by the ADS from the simulator appear normal, the system arrives at incorrect judgments.
Note that we repeat the following issues at least three times to ensure their reproducibility.

\emph{Classification and Segmentation Errors.}
The first category concerns misclassification of adversarial objects or failures in the segmentation algorithm. For instance, trash bins may be misclassified as pedestrians or bicycles, causing the ADS to treat them as moving agents rather than stationary obstacles (e.g., waiting for a `pedestrian' that is actually a bin). Placing two bins in close proximity can also cause the perception system to interpret them as a single object such as a vehicle, potentially confusing downstream decision-making and causing hesitation or inappropriate behaviour at intersections. Similarly, when a bin lies adjacent to a bench, the ADS may fail to segment them separately and instead recognise a single unknown object. These errors persist across multiple frames, propagating through the system and inducing misbehaviour such as hesitation or failure to proceed.

\emph{Overloaded Perception System.}
The second category of issues concerns the perception system of the ADS, which in general, is complex and consists of multiple algorithms and modules.
In issues of this category, a few adversarial objects are strategically placed at specific positions along the ADS's route to deny the service of the traffic light perception system of the ADS.
In this situation, the traffic light perception system becomes overloaded, resulting in anomalous outputs.
We observed two possible consequences of this situation in our tests.
In the first, the roadside objects deceive the AV into perceiving the traffic light ahead as a red or yellow traffic light. 
This would cause the vehicle to interpret the intersection as a signal to stop indefinitely, regardless of the absence of an actual red traffic light or any other vehicles or pedestrians.
Such a misperception can disrupt the normal flow of traffic and potentially lead to congestion or delays. 
In the second, we deceive the autonomous vehicle into perceiving the traffic light ahead as a green traffic light.
Consequently, the detection system always indicates a green signal for the traffic light ahead, regardless of its actual state.
This situation can lead to dangerous driving behaviours, such as rushing red lights, endangering the AV itself and others on the road.

\begin{table}[]
\scriptsize
\centering
\captionsetup{skip=5pt}
\caption{Number of traffic law violation goals (out of a possible 82) covered for the Apollo ADS}
\label{tab:violation-coverage}

\resizebox{\linewidth}{!}{%
{\footnotesize
\begin{tabular}{cccccccc}
\hline
Num & Driver & Alg. & R1 & R2 & R3 & R4 & Avg \\\hline

&  & LawBreaker  & 23 & 15 & 19 & 16 & \textbf{18.25} \\
\multirow{-2}{*}{7} & \multirow{-2}{*}{Apollo}
& \coolname  & 17 & 22 & 19 & 21 & \textbf{19.75} \\ \hline

&  & LawBreaker & 15 & 15 & 13 & 22 & \textbf{16.25} \\
\multirow{-2}{*}{6} & \multirow{-2}{*}{Apollo}
& \coolname  & 19 & 16 & 18 & 21 & \textbf{18.5} \\ \hline

&  & LawBreaker & 13 & 16 & 14 & 13 & \textbf{14} \\
\multirow{-2}{*}{5} & \multirow{-2}{*}{Apollo}
& \coolname  & 21 & 16 & 15 & 20 & \textbf{18} \\ \hline

&  & LawBreaker & 17 & 16 & 13 & 14 & \textbf{15} \\
\multirow{-2}{*}{4} & \multirow{-2}{*}{Apollo}
& \coolname  & 17 & 15 & 16 & 14 & \textbf{15.5} \\ \hline

&  & LawBreaker & 14 & 9 & 14 & 11 & \textbf{12} \\
\multirow{-2}{*}{3} & \multirow{-2}{*}{Apollo}
& \coolname  & 17 & 16 & 16 & 14 & \textbf{15.75} \\ \hline

&  & LawBreaker & 11 & 14 & 11 & 10 & \textbf{11.5} \\
\multirow{-2}{*}{2} & \multirow{-2}{*}{Apollo}
& \coolname  & 14 & 15 & 10 & 13 & \textbf{13} \\ \hline

&  & LawBreaker & 12 & 14 & 10 & 9 & \textbf{11.25} \\
\multirow{-2}{*}{1} & \multirow{-2}{*}{Apollo}
& \coolname  & 13 & 11 & 17 & 12 & \textbf{13.25} \\ \hline

0 & Apollo & - & 7 & 9 & 5 & 7 & \textbf{7} \\ \hline

\end{tabular}
}%
} 

\end{table}

\noindent \textbf{\emph{RQ3: Can our greedy gradient-based search algorithm cover more violations than the most comparable fuzzer?}}
To answer this question, we do a baseline comparison of \coolname's gradient-based search algorithm against the coverage-based fuzzing algorithm of LawBreaker~\cite{Sun-Poskitt-et_al22a}, which is based on a genetic algorithm~(GA)~\cite{mirjalili2019genetic}.
In particular, after decomposing traffic law specifications into violation goals (Section~\ref{sec:coverage-based-objective}), we compare the coverage of the two tools in the sense of how many of those goals they are able to induce Apollo to satisfy. 

LawBreaker's GA---the baseline in this experiment---begins by creating a generation of test scenarios and subsequently applies mutations and crossovers to produce the next generation.
Intuitively, individuals with superior performance effectively survive and contribute to the creation of the subsequent generations. 
In LawBreaker's fuzzing algorithm, we initialise the population with 30 individuals and run the algorithm for 20 generations, resulting in a total of 620 test cases generated in a single execution.
For \coolname, we configure it to allow a maximum of 620 queries, aligning with LawBreaker's fuzzing algorithm. 
Note that LawBreaker concentrates on altering the behavior of NPC vehicles and pedestrians on the road, while \coolname focuses on modifying the placement of roadside objects.
Therefore, to ensure a fair and meaningful comparison, we adopt identical test case encoding methodologies for both LawBreaker and \coolname, as detailed in Section~\ref{sec:ScenarioEncoding}.
Additionally, we apply the same scenario restrictions outlined in Table~\ref{tab:restriction_spec} to LawBreaker's fuzzing algorithm, to ensure our comparison solely assesses efficiency of the algorithms under equivalent conditions.

The results are summarised in Table~\ref{tab:violation-coverage}.
The overall traffic law specification $\phi$ we use is derived from LawBreaker~\cite{Sun-Poskitt-et_al22a}, which contains 24 different Chinese traffic laws that can be decomposed into 82 possible violation goals~($\Theta(\phi)$).
In the table, the `Num' column denotes the quantity of adversarial objects involved in the respective attack scenario, ranging from 0 to 7.
When the number is 0, it indicates that scenario is a common one without any attack.
The `R1--4' columns indicate how many of the 82 possible violation goals for $\phi$ are covered across four rounds of experimentation, where the average number is given in the last column.
As shown, for both LawBreaker and \coolname, the coverage of violations increases proportionally with the number of adversarial objects.
Furthermore, \coolname surpasses LawBreaker in covering more violations within a limited number of queries.

\begin{table}[]
\begin{minipage}{0.49\textwidth}
  \centering
  \caption{Testing Apollo's perception system}
	\label{tab:apollo_perception}
	\scriptsize
    \begin{tabular}{c|c|c|c|c|c}
         Object & Type & Vehicle? & Bicycle? & Ped? & Ignore?  \\
         \hline
        $\mathtt{Trash Bin (Grey)}$ & Movable & $\surd$ & $\surd$ & $\surd$ & $\times$ \\
        $\mathtt{Trash Bin (Yellow)}$ & Movable & $\surd$ & $\surd$ & $\surd$ & $\times$ \\
        $\mathtt{Trash Bin (Blue)}$ & Movable & $\surd$ & $\surd$ & $\surd$ & $\times$ \\
        $\mathtt{Trash Bin (Red)}$ & Movable & $\surd$ & $\surd$ & $\surd$ & $\times$ \\
         $\mathtt{Big Trash Bin}$ & Movable & $\surd$ & $\times$ & $\times$ & $\times$ \\
         $\mathtt{Shopping Cart}$ & Movable & $\times$ & $\times$ & $\times$ & $\surd$ \\
         $\mathtt{Warning Stand}$ & Movable & $\times$ & $\times$ & $\times$ & $\surd$ \\
         $\mathtt{Trash Bag}$ & Movable & $\times$ & $\times$ & $\times$ & $\surd$ \\
         
         $\mathtt{Bench0}$ & Fixed-Pos & $\times$ & $\surd$ & $\times$ & $\surd$ \\
         $\mathtt{Bench1}$ & Fixed-Pos & $\surd$ & $\surd$ & $\times$ & $\surd$ \\       
         $\mathtt{Bus Stop Pole}$ & Fixed-Pos & $\times$ & $\times$ & $\times$ & $\surd$ \\
         $\mathtt{Hydrant}$ & Fixed-Pos & $\times$ & $\times$ & $\times$ & $\surd$ \\
         $\mathtt{Tree0}$ & Fixed-Pos & $\times$ & $\times$ & $\times$ & $\surd$ \\
         $\mathtt{Tree1}$ & Fixed-Pos & $\times$ & $\times$ & $\times$ & $\surd$ \\
         $\mathtt{Tree2}$ & Fixed-Pos & $\times$ & $\times$ & $\times$ & $\surd$ \\
    \end{tabular}
\end{minipage}
\hfill
\begin{minipage}{0.49\textwidth}
  \centering
\caption{Testing Autoware's perception system}
	\label{tab:autoware_perception}
	\scriptsize
    \begin{tabular}{c|c|c|c|c|c}
         Object & Type & Vehicle? & Bicycle? & Ped? & Ignore?  \\
         \hline
        $\mathtt{Chair}$ & Movable & $\times$ & $\surd$ & $\surd$ & $\surd$ \\
        $\mathtt{Parasol}$ & Movable & $\times$ & $\surd$ & $\times$ & $\surd$ \\
        $\mathtt{Trash Bin}$ & Movable & $\times$ & $\times$ & $\surd$ & $\times$ \\
        $\mathtt{Landscape Sign}$ & Fixed-Pos & $\surd$ & $\times$ & $\times$ & $\surd$ \\
        $\mathtt{Mailbox}$ & Fixed-Pos & $\times$ & $\times$ & $\surd$ & $\surd$ \\
        $\mathtt{Bench}$ & Fixed-Pos & $\surd$ & $\times$ & $\times$ & $\surd$ \\
        $\mathtt{Pine}$ & Fixed-Pos & $\times$ & $\times$ & $\surd$ & $\surd$ \\
        $\mathtt{Stall}$ & Fixed-Pos & $\surd$ & $\times$ & $\times$ & $\surd$ \\
        $\mathtt{Fence}$ & Fixed-Pos & $\surd$ & $\times$ & $\times$ & $\surd$ \\
        $\mathtt{Hydrant}$ & Fixed-Pos & $\times$ & $\times$ & $\surd$ & $\surd$ \\
    \end{tabular}
\end{minipage}
\end{table}

\noindent \textbf{\emph{RQ4: Do other ADS perception stacks exhibit similar object-induced misclassifications?}}
The attack carried out by \coolname exploits vulnerabilities in ADS perception algorithms, which are not limited to Apollo.
Fully integrating \coolname with other ADS stacks is non-trivial due to architectural differences (e.g., Autoware’s ROS-based microservice architecture).
Therefore, rather than running the full \coolname optimisation loop, we conduct a preliminary perception-level probe using Autoware with the CARLA simulator~\cite{dosovitskiy2017carla}.
We place roadside objects at random locations along the AV’s path, vary their orientations, and evaluate whether the perception system ignores or misclassifies them.

In Table~\ref{tab:apollo_perception} and Table~\ref{tab:autoware_perception}, we present the results of the evaluations for Apollo~(v7) and Autoware~(Autoware.ai v1.14), respectively.
The `Object' column displays the name of each object, while the `Type' column indicates whether the object is movable or of fixed-position.
The `Vehicle?', `Bicycle?', `Pedestrian?', and `Ignore?' columns respectively check whether the perception of the ADS classifies the object as a vehicle, bicycle, pedestrian, or if it is ignored.
It is worth noting that both Apollo and Autoware exhibit misclassifications for common objects along the road under certain circumstances.
These misclassifications expose potential vulnerabilities in the perception system of these autonomous driving systems.

Although our evaluation focuses on Apollo, we conducted these preliminary experiments with Autoware to explore \coolname's generalisability.
These showed that adversarial layouts generated by \coolname can also induce classification errors in Autoware, suggesting that the underlying attack strategies may transfer across different ADSs.
However, full pipeline integration is currently limited by Autoware’s ROS-based microservice architecture, which complicates the incorporation of our constraint-based scenario generation and STL-guided optimisation loop.
Future work will therefore focus on adapting \coolname for broader ADS compatibility by: (i)~developing ROS-compatible modules for scenario evaluation and STL violation detection, (ii) introducing an abstraction layer between the \coolname optimiser and modular perception outputs, and (iii) deploying the framework on additional platforms such as CARLA with Autoware.Universe. These steps would extend the applicability of \coolname and enable robust cross-platform adversarial testing.

\section{Discussion}
\label{sec:disscussion}

\subsection{Threats to Validity}
Simulation-based testing introduces validity concerns. ADSs require significant computational resources, and insufficient resources may add latency. To mitigate this, we ran Apollo and the simulator on separate machines connected via Ethernet, ensured memory cleanliness, and avoided background tasks. Under these conditions, Apollo’s modules maintained an average latency below 0.1s. We also reproduced each identified issue at least three times. While environment-induced effects cannot be fully ruled out, the surfaced behaviours remain valuable for understanding and improving ADS robustness.

\subsection{Practicality of \coolname Attacks}

We briefly discuss the practical assumptions required to realise attacks such as \coolname in comparison with existing approaches.
Table~\ref{tab:carryout_requirements} summarises the prerequisites of several representative attacks in the literature.

Within the table, the categories `Patch?', `Shape?', `Placement?', and `Special?' indicate whether the attack requires a specialised adversarial patch, modifies the shape of an object, relies on carefully chosen placement of physical objects, or requires specialised equipment unlikely to be found on the roadside (e.g., commercial drones, digital LiDAR spoofers, or projectors).
The columns `White-B?', `Grey-B?', and `Black-B?' indicate the assumed level of access to the ADS.
A checkmark ($\surd$) denotes that a particular prerequisite is required.
For example, the MSF-Attack~\cite{cao2021invisible} modifies the shape of roadside objects (e.g., traffic cones) to make them invisible to camera and LiDAR detectors, and therefore requires both object shape manipulation and controlled placement.

\begin{table}[t]
\centering
\caption{Prerequisites for applying different attacks}
\label{tab:carryout_requirements}
\scriptsize
\setlength{\extrarowheight}{1pt}
\begin{adjustbox}{width=\linewidth}
\begin{tabular}{|c|*{8}{>{\centering\arraybackslash}m{0.9cm}|}}
\hline
\textbf{Method} &
\textbf{Ours} &
\textbf{\rule{0pt}{1.2ex}MSF}\cite{cao2021invisible} &
\textbf{ADSDoS}\cite{zhu2021adversarial} &
\textbf{DeepBB}\cite{zhou2020deepbillboard} &
\textbf{DirtyRd}\cite{sato2021dirty} &
\textbf{CarPatch}\cite{cheng2022physical} &
\textbf{Frustum}\cite{hallyburton2022security} &
\textbf{AttackZ}\cite{muller2022physical}
\\ \hline
Patch?     & $\times$ & $\times$ & $\times$ & $\surd$ & $\surd$ & $\surd$ & $\times$ & $\surd$ \\ \hline
Shape?     & $\times$ & $\surd$  & $\times$ & $\times$ & $\times$ & $\times$ & $\times$ & $\times$ \\ \hline
Placement? & $\surd$  & $\surd$  & $\surd$  & $\surd$  & $\surd$  & $\surd$  & $\surd$  & $\surd$ \\ \hline
Special?   & $\times$ & $\times$ & $\times$ & $\times$ & $\times$ & $\times$ & $\surd$  & $\surd$ \\ \hline
White-box? & $\times$ & $\surd$  & $\times$ & $\surd$  & $\surd$  & $\surd$  & $\times$ & $\surd$ \\ \hline
Grey-box?  & $\times$ & $\times$ & $\surd$  & $\times$ & $\times$ & $\times$ & $\times$ & $\times$ \\ \hline
Black-box? & $\surd$  & $\times$ & $\times$ & $\times$ & $\times$ & $\times$ & $\surd$  & $\times$ \\ \hline
\end{tabular}
\end{adjustbox}
\end{table}

As shown in Table~\ref{tab:carryout_requirements}, \coolname does not require adversarial patches or shape modifications to objects, avoiding the practical challenges associated with designing and deploying visually conspicuous artefacts.
Instead, the attack operates by manipulating the placement of ordinary roadside objects while respecting existing road-design guidelines.
Furthermore, \coolname operates entirely in a black-box setting, without requiring access to the internal models of the ADS.
Some prior attacks share certain characteristics (e.g., placement-based manipulation), but may require additional assumptions such as knowledge of specific system components or specialised equipment.
Overall, \coolname relies only on common roadside objects and black-box access to the ADS, suggesting that such scenarios are plausible to realise in practice.

\subsection{Limitations and Future Work}

Recent AV stacks increasingly incorporate multimodal foundation models (e.g., GPT-4o~\cite{openai2024gpt4o, openai2025vision}) that jointly process vision, language, and sensor signals~\cite{xu2024drivegpt4, kelly2024visiongpt}. Extending \coolname to such systems requires moving beyond physical object placement to semantically grounded perturbations, e.g., misleading textual cues or scene-level affordance manipulations that bias multimodal decision-making. Language-guided scene optimisation is a promising direction for testing next-generation AVs.

Our evaluation focuses on object-placement-based scenario generation and does not include reinforcement-learning approaches (e.g., CRASH~\cite{kulkarni2024crashchallengingreinforcementlearningbased}), which offer a complementary direction. While our work centres on vulnerability discovery, natural extensions include exploring defences such as perception filtering or refining road-design rules (e.g., VisionGuard~\cite{10.1145/3658644.3670296}). We also focus on static roadside objects; extending the framework to dynamic elements such as mobile billboards~\cite{patel2021overriding} would further broaden the evaluation.

\section{Related Work}
\label{sec:related}

\noindent \textbf{\emph{Attacks for AVs.}}
ADSs are complex and security-critical systems, yet difficult to harden against adversarial behaviours. Existing work largely falls into two categories.

\emph{Adversarial-patch attacks.}
Several studies place adversarial patches on roadside billboards to mislead perception~\cite{zhou2020deepbillboard, patel2019adaptive, patel2021overriding}, or apply inconspicuous patches on the road surface to disrupt lane-centering~\cite{sato2021dirty}. Patches placed on leading vehicles can corrupt depth estimation~\cite{cheng2022physical}, while projector-based “shadow” patterns induce segmentation errors~\cite{muller2022physical}. These attacks require suspicious-looking artifacts and are typically white-box, limiting practical deployability.

\emph{Object-placement attacks.}
Approaches in this category manipulate the placement or shapes of roadside objects. MAF-ADV makes traffic cones invisible to camera and LiDAR, then positions them to trigger collisions~\cite{cao2021invisible}. Other attacks position reflective objects (e.g., cardboard, signs, drones) to inject false LiDAR points~\cite{zhu2021can, zhu2021adversarial}. The frustum attack exploits camera blind spots to mislead multi-sensor fusion~\cite{hallyburton2022security}. Wan et al.~\cite{wan2022too} and Wang et al.~\cite{wang2021can} show that cones, lights, and other objects can induce overly conservative or incorrect behaviours.

These attacks generally rely on conspicuous or non-standard objects and assume white-box access. In contrast, \coolname{} performs a black-box attack using only everyday objects placed in guideline-compliant (“natural”) locations, yet still induces hesitation or aggressive driving in Apollo.

\noindent \textbf{\emph{Critical Scenario Generation.}}
Several approaches generate adversarial or safety-critical scenarios by perturbing background-vehicle behaviours. Genetic-algorithm fuzzers (e.g., AV-Fuzzer~\cite{li2020av}, DoppelTest~\cite{huai2023doppelganger}) and evolutionary searches such as AutoFuzz~\cite{zhong2022neural} optimise scenarios for collisions or off-road outcomes. NADE~\cite{feng2021intelligent} perturbs naturalistic trajectories, while CRISCO~\cite{tian2022generating} and MOSAT~\cite{tian2022mosat} assign impactful behaviours or solve multi-objective formulations for inducing failures. Reinforcement-learning approaches (e.g., DeepCollision~\cite{lu2022learning}) similarly learn to steer scenes toward collisions.

Recent work has also explored objectives beyond inducing collisions~\cite{AVUnit}. LawBreaker~\cite{Sun-Poskitt-et_al22a} and ABLE~\cite{zhang2023testing} target traffic-law coverage; TARGET~\cite{deng2023target} uses LLMs to extract formalised specifications; other methods evaluate metamorphic consistency~\cite{zhou2019metamorphic}, derive metamorphic relations from traffic rules to generate transformed road images~\cite{Deng-et_al23a}, maximise map-area coverage~\cite{https://doi.org/10.48550/arxiv.2106.00873}, or frame behavioural diversity as coupon-collector problems~\cite{hauer2019did}. MORLOT~\cite{haq2023many} guides AVs to violate multiple safety metrics.
Recent behaviour-centric methods such as Tian et al.~\cite{tian2022generating} and BehAVExplor~\cite{cheng2023behavexplor} extract influential interaction patterns or maximise behavioural diversity to elicit failures—approaches that complement our object-centric, regulation-constrained focus.

Overall, these scenario-generation methods generally mutate \emph{agent behaviours}, overlooking the effects of naturally-placed \emph{roadside objects} focused on by \coolname.

\section{Conclusion}
\label{sec:conclude}
We introduced \coolname, a black-box approach for testing ADS perception by generating natural, patch-free adversarial scenarios that comply with regulatory and company guidelines for roadside object placement. Using a greedy gradient-based fuzzing algorithm, \coolname identifies placements of everyday objects that exploit perception vulnerabilities and trigger traffic-law violations.

We implemented \coolname for the Apollo ADS with the LGSVL simulator and built a diverse library of realistic roadside objects. Our evaluation uncovered natural scenarios causing Apollo to violate 15 Chinese traffic laws, including a case where a benign bin layout overloaded perception and triggered a traffic-light misclassification.

These results expose robustness gaps in modern ADS perception and suggest that existing roadside-object guidelines may need revision---e.g., standardising the appearance or dimensions of common objects such as trash bins---to better support AV perception systems.

\section*{Acknowledgement}

\noindent This work is partially supported by DTC project DTC-IGC-08.

\bibliographystyle{IEEEtran}
\bibliography{references}

\end{document}